# Enhanced version of AdaBoostM1 with J48 Tree learning method

Kyongche Kang
Carnegie Mellon University
5000 Forbes Avenue
Pittsburgh, PA 15213
kyongchk@andrew.cmu.edu

Jack Michalak
Carnegie Mellon University
5000 Forbes Avenue
Pittsburgh, PA 15213
jmichala@andrew.cmu.edu

## ABSTRACT
Machine Learning focuses on the construction and study of systems that can learn from data. This is connected with the classification problem, which usually is what Machine Learning algorithms are designed to solve. When a machine learning method is used by people with no special expertise in machine learning, it is important that the method be 'robust' in classification, in the sense that reasonable performance is obtained with minimal tuning of the problem at hand. Algorithms are evaluated based on how 'robust' they can classify the given data. In this paper, we propose a quantifiable measure of 'robustness', and describe a particular learning method that is robust according to this measure in the context of classification problem. We proposed Adaptive Boosting (AdaBoostM1) with J48(C4.5 tree) as a base learner with tuning weight threshold (P) and number of iterations (I) for boosting algorithm. To benchmark the performance, we used the baseline classifier, AdaBoostM1 with Decision Stump as base learner without tuning parameters. By tuning parameters and using J48 as base learner, we are able to reduce the overall average error rate ratio (errorC/errorNB) from 2.4 to 0.9 for development sets of data and 2.1 to 1.2 for evaluation sets of data.

## 1. INTRODUCTION
Machine Learning has become a very large field in our society. Large datasets abound as technology has allowed us to collect more and more data, and so the ability to use computers to extract patterns from these data has become increasingly important. But we would hope that such technological advancements would not be relegated to academia and scientists, but rather that they could be used by non-specialists as well. It is for this reason that there is an interest in the robustness of a classifier: by finding an algorithm that performs reasonably well in all situations, we can provide it as a standard for non-specalists to use with minimal concern.

In this paper, we define robustness as a measure to be the error associated with a given classifier divided by the error associated with the Naive Bayes classifier on the same dataset. We begin with Adaptive Boosting (AdaBoost), a boosting classifier, and tune it with the qualities of robustness in mind. For each parameter of the classifier, we train and test it on our datasets using cross-validation to determine the best possible values. For some parameters with finitely many discrete values, we tested all of their possible values, while with others we chose an increment to cover all of the reasonable values for the parameter. After we selected the parameters using this methodology, we tested our tuned classifier against the baseline classifier we started with on unseen test sets to determine how much the classifier had improved in robustness.

For both the development set and the evaluation set, we saw an increase in the robustness of this tuned classifier over the baseline classifier. Unsurprisingly, the test set had a smaller gap than the development set, but it is good to see that most of the improvement is independent of what set the classifier is run on, and shows that our classifier is indeed robust.

## 2. BACKGROUND
Our proposal extends AdaBoost, a boosting algorithm, for classification. Our desire is to create a more robust classifier using the tools in Machine Learning that are available to us. We have taken advantage of the boosting paradigm via Weka's AdaBoostM1 class (weka.classifiers.meta.AdaBoostM1) and have applied it to decision trees with Weka's J48 class (weka.classifiers.trees.J48).

Boosting involves the process of training a weak classifier on the datasets iteratively: the first time it is trained normally, but then the training examples have different weights on each following iteration. The differing weights allows later iterations to focus harder on the portions of the training set that previous iterations have misclassified, so that when they are combined they can have a much better classification ability when used together. Boosting also typically weighs the different trained classifiers with an entirely separate set of weights according to their accuracy across the training set. Boosting algorithm is an example of ensemble methods, which are learning algorithms that construct a set of classifiers and then classify new data points by taking a (weighted) vote of their prediction. [6] It in a sense is a special case of the ensemble model within Machine Learning: it can be seen as an ensemble classifier in which all of the models within it are of the same algorithm.

Boosting was originally introduced by Schapire [5]. Researchers were interested in if the existence of weak learners (ie, those doing only slightly above pure chance) implied the existence of strong learners (those doing arbitrarily well on a learning problem). Boosting was Schapire's answer to the problem; it is a constructive proof that the answer is yes. Boosting allows one to repeatedly train the same weak learner, and the result at the end is a strong learner. Schapire would later improve his algorithm by turning it into Adaptive Boosting, which made the learning more efficient via weighting, and is the more popular boosting algorithm today, though many other boosting algorithms have since been introduced. [1]

Decision Trees are a very large class of learning algorithms. The general idea in a Decision Tree is that when classifying, the classifier will begin at the root of the tree and then move from node to node according to the value of the feature at that node for the specific instance. For example, if the root node attribute is weather, then the branches may correspond to values of *Sunny*, *Windy*, and *Rainy*. The classifier follows this procedure according to the attributes of the instance until it reaches a leaf node, which indicates what decision should be made for that particular instance (which class should be predicted).

In order to generate this tree model as a learner, it needs to have some idea of what attribute is most beneficial in the classification process, and for this purpose the idea of "information gain" was created. Basically this means that by choosing to branch on a particular attribute, the classifier would be more likely to guess the correct class given the probabilities -- the classifier should have reduced the overall entropy of the data by branching (entropy has a formal definition, but it will suffice to think of it as randomness for now) by as much as possible based on its choice of attribute. A decision tree algorithm will repeatedly choose the attribute which causes the greatest decrease in entropy in order to ensure classification can be made quickly. Eventually the algorithm will either get to a point in the branching where either all of the training examples have the same class, or otherwise it has run out of attributes. If it is the former, the algorithm can just make this a decision leaf with the appropriate class label; if the latter, then the majority class of the remaining training examples must be chosen for the best result.

J48 is actually a Java implementation of the C4.5 algorithm introduced by Quinlan. [4] His algorithm provided some nice improvements from his previous ID3, including allowing continuous variables to be features by choosing a threshold value at which to split, making it more versatile, and doing what is called a pruning of the tree, which makes the actual classification tasks more efficient. Basically the algorithm will filter back through the entire tree after it has been created, looking for branches that aren't particularly helpful according to some input parameter, and replacing them with leaf nodes if they don't surpass the threshold.

## 3. PROPOSED LEARNING METHOD

We propose Adaptive Boosting classifier with J48 as a base decision tree. The weka name is AdaBoostM1. As mentioned above, AdaBoost is a meta algorithm that can be used in conjunction with many other learning algorithms to improve their performance. It works by repeatedly running a given 'weak' or base learning algorithm on various distributions over the training data, and then combining the classifiers produced by the weak learner into a single composite classifier. [1] We will refer to our proposed learning method as L5. With this classifier, we tuned a number of parameters to optimize the performance. The following parameters are tuned and their impacts on the performance are analyzed.

For all the learning, we applied 10-fold cross-validation. By doing so, we are able to avoid overfitting and get more accurate results on testing sets and generalize predictive errors.

### 3.1 Parameters for AdaBoostM1

*Weight Threshold (P)*

This refers to the percentage of weight mass to base training on. As mentioned above, each training example is given weights when combined. This allows later iterations to focus harder on the portions of the training set that previous iterations have misclassified. The default value is 100, which essentially means no weight. We tried varying it from 10 to 100, incrementing by 10. Our classifier seemed to be most sensitive to this parameter.

*Number of Iterations (I)*

This refers to the number of iterations that AdaBoostM1 calls for Weak Classifier, which is J48 in this case. The default value is 10. We tried varying it from 10 to 50, incrementing by 10. Our learning method seemed to be sensitive to tuning this parameter.

*Resampling for boosting (Q)*

This refers to whether the boosting algorithm uses resampling of the data or not. We find that resampling of the data is almost always better in all cases. Resampling method is clearly the preferred method in boosting algorithm. [2]

*Base Classifier (W)*

This refers to the base classifier that AdaBoostM1 uses as a base learner. The default is Decision Stump tree learning. In our learning method, we used J48.

### 3.2 Parameters for J48

*Pruning Confidence (C)*

Pruning the tree helps to boost performance by removing branches which are not helpful. This prevents overfitting to certain extent. Pruning confidence is used to calculate an upper bound on error rate at leaf. The default value is 0.25. We tried varying it from 0.1 to 0.5, incrementing by 0.05. We see that the classifier was not particularly sensitive to change in this parameter.

*Minimum number of instances (M)*

This refers to the number of instances per node in fitting the tree. The default is 2 and we tried varying it from 1 to 4 incrementing by 1. Our J48 tree was not particularly sensitive to this parameter.

*Number of folds (N)*

This parameter refers to a number of folds for reduced error pruning. One fold is used as pruning set. The default value is 3 and we tried varying it from 1 to 5, incrementing by 1. We see that our classifier is not particularly sensitive to tuning this parameter.

In summary, we selected weight threshold (P) and number of iterations (I) for automatic tuning. With CVParameterSelection function in weka (weka.classifiers.meta.CVParameterSelection), we are able to implement 10-fold cross-validation as well as automatic tuning that allows the learning method to tune the parameters for best performance by each dataset. Then we fix resampling for boosting (Q) to be used for fast evaluation and Base Classifier (W) to be J48.

In order to benchmark the performance of L5, we used a baseline classifier, Adaptive Boosting with Decision Stump as a base decision tree. This is not going to tune any of the parameters as we are interested in the impact of tuning parameters on performance. The base classifier will use the default settings for all the parameters. We will call this LB. The results of our experiment are the following.

## 4. EXPERIMENTAL RESULTS

We are given two types of datasets - development and evaluation. Development set consists of 24 datasets with the true labels provided. In evaluation set, we are given 20 datasets without the true labels. For choosing the learning method, we used the development datasets to train the learning method and the baseline classifier, and parameters to tune to develop our final learning method. Using the evaluation datasets, we are able to benchmark its performance against different learning methods as well as against the baseline classifier, LB.

In this experiment, we used the error rate ratios of error rate of our classifier and error rate of Naive Bayes classifier (errorC/errorNB). It is widely accepted that Naive Bayes classifier is a strong classifier, which works universally well in most of the datasets. Rather than giving error rates of the learning method alone, by using the ratios, we are able to benchmark our learning method's performances compared to Naive Bayes classifier and give clearer picture of how well it performs. In addition, comparing with Naive Bayes also provides comparison of single classifier against ensemble method classifier, which incorporates more than one learning algorithm to improve performance.

The below tables show the summary of each dataset in development and evaluation sets, and also the error rates and errorC/errorNB ratios for LB and L5 learners for development datasets. In almost all cases, L5 was able to outperform LB. LB did better than L5 in only a couple of datasets, but there is enough evidence to believe that L5 with tuning parameters is always better than LB, with untuned parameters.

The above table summarizes the performances for LB and L5 for development and evaluation datasets. Baseline classifier (LB) had average rates of errorC/errorNB around the similar level, above 2.0, for both development and evaluation sets. LB was slightly worse for development sets, but this could be accounted for the fact that development comprised more datasets (24) than evaluation (20) did . Our tuned classifier, L5, performed around the same level for both development and evaluation datasets as well, with the average rates of errorC/errorNB of 0.9 for development and 1.2 for evaluation sets. L5 did slightly better in development sets than in evaluation sets and this seems in contrast with what we observed earlier with LB.

Table 3.

|  | Baseline Learner (LB) | Tuned Learner (L5) |
|---|---|---|
| Development | 2.4 | 0.9 |
| Evaluation | 2.1 | 1.2 |

**Table 1.**

| Name | No. of training/ testing | No. features | ErrorC/ErrorNB (Error rate) LB | ErrorC/ErrorNB (Error rate) L5 |
|---|---|---|---|---|
| anneal | 602/296 | 38 | 2.71 (0.16) | 0.29 (0.02) |
| audiology | 151/75 | 69 | 1.68 (0.49) | 0.55 (0.16) |
| autos | 137/68 | 25 | 1.08 (0.60) | 0.95 (0.53) |
| balance-scale | 419/206 | 4 | 1.17 (0.37) | 0.83 (0.27) |
| breast-cancer | 192/94 | 9 | 0.97 (0.30) | 0.83 (0.26) |
| colic | 247/121 | 22 | 0.96 (0.18) | 0.83 (0.16) |
| credit-a | 462/228 | 15 | 0.70 (0.26) | 0.73 (0.27) |
| diabetes | 515/253 | 8 | 1.16 (0.28) | 1.24 (0.30) |
| glass | 143/71 | 9 | 0.82 (0.59) | 0.51 (0.37) |
| heart-c | 203/100 | 13 | 0.67 (0.12) | 0.89 (0.16) |
| hepatitis | 104/51 | 19 | 1.0 (0.12) | 0.83 (0.10) |
| hypothyroid2 | 2527/1245 | 27 | 1.38 (0.08) | 0.08 (0.005) |
| ionosphere | 235/116 | 34 | 1.38 (0.19) | 0.81 (0.11) |
| labor | 38/19 | 16 | 5.0 (0.26) | 2.0 (0.11) |
| lymph | 99/49 | 18 | 0.9 (0.18) | 0.80 (0.16) |
| mushroom | 5443/2681 | 22 | 2.31 (0.30) | 0.30 (2.31) |
| segment | 1548/762 | 19 | 3.89 (0.73) | 0.17 (0.03) |
| sonar | 139/69 | 60 | 0.95 (0.54) | 1.36 (0.76) |
| soybean | 458/225 | 35 | 12.0 (0.85) | 1.19 (0.08) |
| splice | 2137/1053 | 61 | 1.02 (0.76) | 0.99 (0.74) |
| vehicle | 567/279 | 18 | 1.21 (0.60) | 0.44 (0.22) |
| vote | 291/144 | 16 | 0.17 (0.01) | 0.25 (0.02) |
| vowel | 663/327 | 13 | 1.82 (0.87) | 0.98 (0.47) |
| zoo | 68/33 | 17 | 12.0 (0.36) | 1.0 (0.03) |

**Table2.**

| Name | No. of training / testing | No. of features |
|---|---|---|
| arrhythmia | 303 / 149 | 279 |
| breast-w | 468 / 231 | 9 |
| car | 1158 / 570 | 6 |
| cmc | 987 / 486 | 10 |
| credit-g | 670 / 330 | 20 |
| cylinder-bands | 362 / 178 | 39 |
| dermatology | 245 / 121 | 34 |
| ecoli | 225 / 111 | 7 |
| flags | 130 / 64 | 30 |
| haberman | 205 / 101 | 3 |
| heart-h | 203 / 97 | 14 |
| heart-statlog | 181 / 89 | 13 |
| kr-vs-kp | 2141 / 1055 | 36 |
| liver-disorders | 231 / 114 | 6 |
| mfeat-factors | 1340 / 660 | 216 |
| mfeat-fourier | 1340 / 660 | 76 |
| mfeat-karhunen | 1340 / 660 | 64 |
| primary-tumor | 227 / 112 | 17 |
| sick | 2527 / 1245 | 29 |
| sonar | 139 / 69 | 60 |

## 5. CONCLUSIONS

In this paper, we explored the best learning method that can be generalized to classify different sets of data. We experimented how this learning method performed by tuning different parameters, benchmarked its performance with baseline learner and trained and tested on different datasets. We proposed Adaptive Boosting algorithm (weka name: AdaBoostM1) with J48 decision tree base. To further develop our learning method, we optimized automatic tuning of weight threshold (P), number of iterations (I), using resampling for boosting (Q) for AdaBoostM1 and J48 decision tree as base. The baseline classifier is untuned AdaBoostM1 with Decision Stump as base. The construction of this algorithm was done on 24 development datasets and tested on 20 evaluation datasets. This produced the average error rate errorC/errorNB of 0.9 and 1.2 respectively. Compared with baseline classifier, 2.4 and 2.1 respectively, our learning method seemed to perform well.

We attribute this performance to the fact that Adaptive Boosting is an ensemble method. As we have seen, ensemble method generally seems to further improve predictive performance than single learning method. Through our experiments, we conclude that our proposed method is robust in classification with minimal tuning of the parameters. Hence, ensemble method is superior and we were able to produce robust performance for classification problems with AdaBoost algorithm.

## 6. APPENDIX

Throughout the project, we split the work evenly. We always met up for project milestones and brainstormed ideas, implemented and turned them into deliverables. We all wrote codes, discussed and selected classifiers and wrote reports. This also applied to Milestone 6. We split the sections to write individually and compiled our writings and went over with each other to edit and finalize it.

We would like to thank Professor Cohen and Xing as well as TAs for making this course wonderful. Although there were some miscommunications, we believe that the course has been fantastic and challenging enough for us to learn more about Machine Learning. To give some feedback on the project, in the future, we would like this to incorporate data analysis aspect, where students use classifiers and Machine Learning algorithms to solve some of the real problems. This may be challenging with given number of students taking this course, so perhaps it can be guided data analysis, similar to project milestones.